\documentclass[10pt,twocolumn,letterpaper]{article}

\usepackage{wacv}
\usepackage{times}
\usepackage{epsfig}
\usepackage{graphicx}
\usepackage{amsmath}
\usepackage{amssymb}
\usepackage{booktabs}
\usepackage{multirow}
\usepackage{dblfloatfix}
\usepackage{array}
\newcolumntype{C}[1]{>{\centering}p{#1}}
\usepackage{makecell}

\def\wacvPaperID{495} 

\wacvfinalcopy 

\ifwacvfinal
\usepackage[breaklinks=true,bookmarks=false]{hyperref}
\else
\usepackage[pagebackref=true,breaklinks=true,colorlinks,bookmarks=false]{hyperref}
\fi

\ifwacvfinal
\fi

\begin{document}

\title{TB-Net: A Three-Stream Boundary-Aware Network for Fine-Grained Pavement Disease Segmentation
}

\author{Yujia Zhang\textsuperscript{\rm 1}, Qianzhong Li\textsuperscript{\rm 1,\rm 2}, Xiaoguang Zhao\textsuperscript{\rm 1}, and Min Tan\textsuperscript{\rm 1}\\
\textsuperscript{\rm 1}Institute of Automation, Chinese Academy of Sciences\\
\textsuperscript{\rm 2}School of Artificial Intelligence, University of Chinese Academy of Sciences\\\
\{zhangyujia2014, liqianzhong2017, xiaoguang.zhao, min.tan\}@ia.ac.cn}

\maketitle
\thispagestyle{empty}

\begin{abstract}
   Regular pavement inspection plays a significant role in road maintenance for safety assurance. Existing methods mainly address the tasks of crack detection and segmentation that are only tailored for long-thin crack disease. However, there are many other types of diseases with a wider variety of sizes and patterns that are also essential to segment in practice, bringing more challenges towards fine-grained pavement inspection. In this paper, our goal is not only to automatically segment cracks, but also to segment other complex pavement diseases as well as typical landmarks (markings, runway lights, etc.) and commonly seen water/oil stains in a single model. To this end, we propose a three-stream boundary-aware network (TB-Net). It consists of three streams fusing the low-level spatial and the high-level contextual representations as well as the detailed boundary information. Specifically, the spatial stream captures rich spatial features. The context stream, where an attention mechanism is utilized, models the contextual relationships over local features. The boundary stream learns detailed boundaries using a global-gated convolution to further refine the segmentation outputs. The network is trained using a dual-task loss in an end-to-end manner, and experiments on a newly collected fine-grained pavement disease dataset show the effectiveness of our TB-Net.
\end{abstract}

\section{Introduction}
Pavement disease segmentation is a fundamental problem in maintaining the condition of transport infrastructure such as airports, bridges, and roads~\cite{fan2019road,zhu2020weakly}. Previous works often only focus on long-thin cracks and address the task of crack segmentation~\cite{fang2020novel,li2018automatic,xie2020main} using some existing crack datasets (e.g., AigleRN~\cite{shi2016automatic}, CrackForest~\cite{chambon2011automatic}, DeepCrack~\cite{liu2019deepcrack}), where the size of cracks tends to be large leading to easier segmentation. However, in practice, other pavement diseases are also critical to segment in order to enhance public safety. This inevitably poses more challenges due to the irregular shapes and various sizes of diseases, some of which may even be difficult to recognize by humans. Therefore, it is of great significance to develop a fine-grained segmentation technique, which can automatically detect different pavement diseases for detailed inspection and generate pixel-level segmentation outputs.

\begin{figure*}[t]
\centerline{\includegraphics[width=0.88\textwidth]{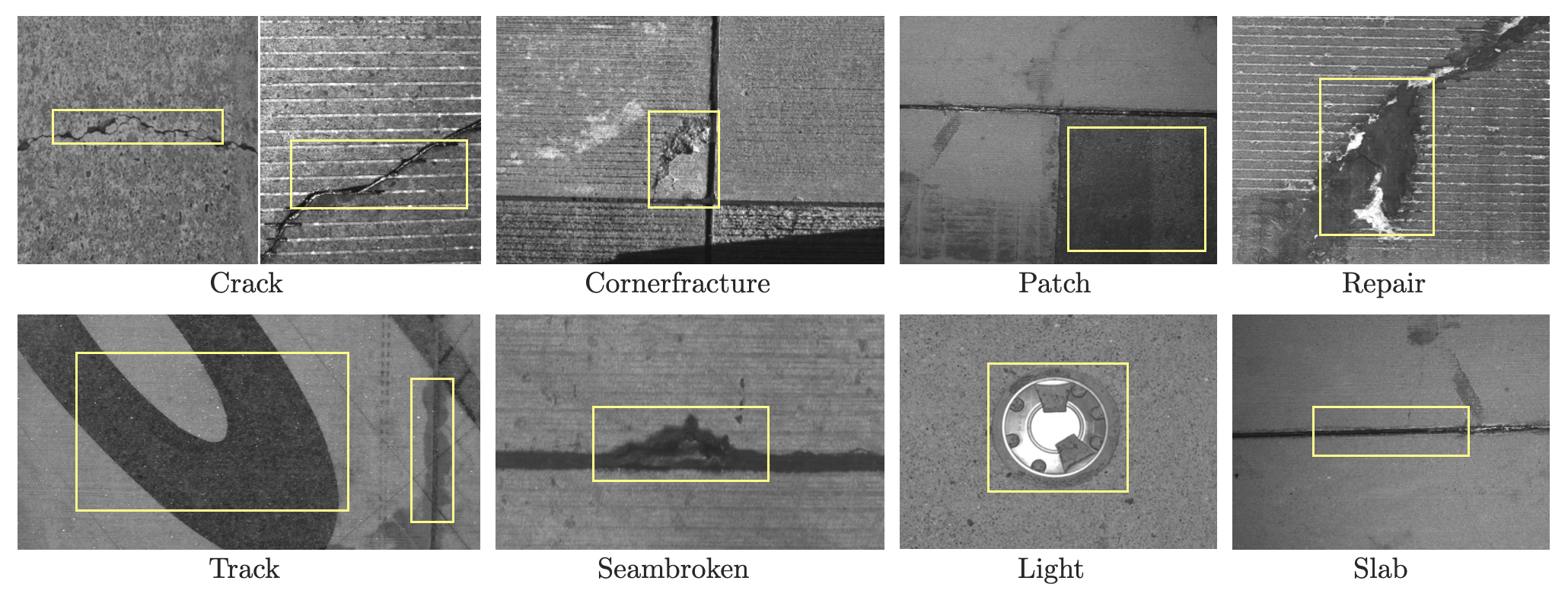}}
\caption{The examples from the newly collected dataset for fine-grained pavement segmentation. They cover different diseases and typical landmarks (markings, runway lights, etc) and commonly seen water/oil stains that can reflect the detailed pavement condition. The yellow rectangles indicate the regions of different diseases (as well as non-diseases). Note that all images are captured with a gray-scale camera.}
\label{fig1}
\end{figure*}

In addition to segmenting different diseases, other typical landmarks (such as markings, runway lights, etc.) as well as commonly seen water/oil stains are also essential to segment for better understanding the general condition of a pavement. Some examples from the newly collected disease dataset for fine-grained pavement inspection are shown in Figure~\ref{fig1}. The yellow rectangles indicate the regions of different diseases (as well as non-diseases). Note that all images are captured with a gray-scale camera.

Specifically, \emph{Crack} has long-thin shapes with different orientation angles (horizontal, vertical and oblique). The small break in the corner of a block is \emph{Cornerfracture}. \emph{Patch} is the rectangle-like cement patch and \emph{Repair} is the strip-shaped asphalt repair. Markings that have ring/line shapes and commonly seen water/oil stains are all classified as \emph{Track}. The small damaged area along the edge of a block is \emph{Seambroken}. \emph{Light} is the ground lighting equipment that is especially deployed at airports. \emph{Slab} is the normal gap between two cement or asphalt blocks. These classes cover different diseases and typical objects which can reflect the detailed pavement condition.

Images of pavement surfaces tend to contain noises due to low contrast, varying illumination, and inconsistent textures. Several previous works on crack detection and segmentation of textured images have been studied to address this challenge by utilizing deep neural networks. For example, Zhang et al.~\cite{zhang2016road} first proposed a framework based on Convolutional Neural Networks (CNNs) that has 10 hidden layers on patch-based images to address the crack detection task. Dung and Anh~\cite{dung2019autonomous} developed an encoder-decoder-based Fully Convolutional Networks (FCNs) to distinguish crack/non-crack pixels. Liu et al.~\cite{liu2019deepcrack} later introduced a hierarchical segmentation network, DeepCrack, that integrates FCNs with Conditional Random Fields (CRFs) and Guided Filtering (GF)~\cite{he2012guided} in order to better exploit the multi-scale and the multi-level features for refinement. Another encoder-decoder architecture built on SegNet~\cite{badrinarayanan2017segnet} for semantic segmentation, is also designed to capture hierarchical convolutional features and perform pixel-wise crack detection~\cite{zou2018deepcrack}. With the recent development of adversarial training, Zhang et al.~\cite{zhang2020crackgan} utilized Generative Adversarial Networks (GANs)~\cite{goodfellow2014generative} to eliminate the influence of noises from large background areas for crack detection.

These above methods aim to tackle the problem of crack detection/segmentation. However, for fine-grained pavement disease segmentation, models trained only on cracks cannot meet the requirement of correctly detecting/segmenting other pavement diseases (as well as non-diseases). On one hand, similarities among some certain classes make it difficult to discriminate. On the other hand, diseases within the same class may possess quite different appearances, such as markings and water/oil stains in \emph{Track}. Moreover, these classes have a large variety of shapes and sizes, which poses more challenges towards fine-grained pavement segmentation.

Recent years have seen great advances in the field of semantic image segmentation. Earlier approaches widely make use of FCNs that take arbitrary-sized inputs to produce pixel-wise predictions~\cite{long2015fully,chen2014semantic}. Since affluent spatial representations are crucial for this dense classification problem, based on fully-convolution-style networks, U-Net~\cite{ronneberger2015u} is further developed by adding skip connections in order to exploit features from the middle layers. A multi-path refinement network, RefineNet~\cite{lin2017refinenet}, is also introduced to capture multi-level features and produce high-resolution segmentation maps. Some other methods are tasked to segment objects at multiple scales by encoding the multi-scale contextual features. For instance, DeepLab v3~\cite{chen2017rethinking} and DeepLab v3+~\cite{chen2018encoder} are proposed by applying the atrous spatial pyramid pooling (ASPP) module~\cite{chen2017deeplab}, which is built upon atrous convolution. They aim to capture the multi-scale contexts with the filters at different atrous rates. DenseASPP~\cite{yang2018denseaspp} further boosts the segmentation performance by densely connecting multiple atrous convolutional layers in order to achieve a larger receptive field.

\begin{figure*}[b]
\centerline{\includegraphics[width=0.9\textwidth]{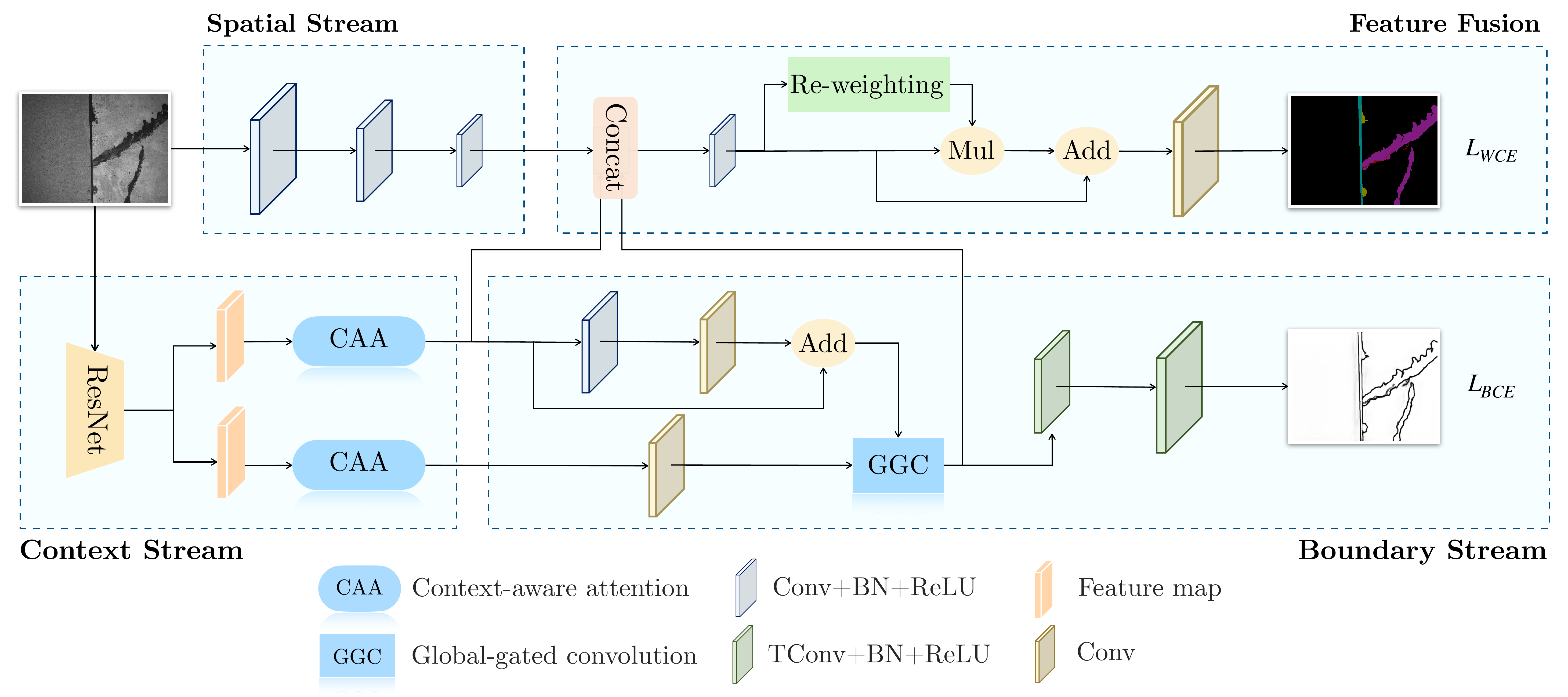}}
\caption{The details of the TB-Net architecture. The three streams are proposed to take full advantages of the low-level spatial, the high-level contextual and the detailed boundary information. Context-aware attention modules are developed in the context stream to model a wide range of contextual relationships. The boundary stream with the introduced global-gated convolution provides additional regularization to further refine the segmentation results. The three learned features are then fused to produce the final segmentation results.}
\label{fig2}
\end{figure*}

More recently, some other effective segmentation networks are designed by fusing different information to form dense image representations. For example, Yu et al.~\cite{yu2018bisenet} proposed BiSeNet which fuses the spatial information and the global contexts. This real-time model achieves higher results against other non-real-time algorithms on popular benchmarks. Takikawa et al.~\cite{takikawa2019gated} introduced Gated-SCNN that processes shape information in parallel to the regular spatial stream. Their model can produce sharper predictions around boundaries, especially on thin and small objects.

Inspired by the recent works that benefit from fusing different information, we propose a three-stream boundary-aware network (TB-Net) for the task of fine-grained pavement disease segmentation, which can explicitly exploit the spatial, the contextual and the boundary information. Specifically, it consists of three streams: 1) The spatial stream aims to obtain high-resolution spatial information. 2) The context stream utilizes an attention mechanism that models the global dependencies. 3) The boundary stream learns boundary features using a global-gated convolution. The features of these three streams are fused to produce the final pixel-level predictions. The network is trained using a dual-task loss to regularize both the segmentation and the boundary outputs. We evaluate the proposed method on a newly collected dataset, and the detailed analysis demonstrates the effectiveness of our TB-Net.

The contributions of this work are as follows:
\begin{itemize}
    \item We propose a three-stream boundary-aware network that takes full advantages of the low-level spatial, the high-level contextual and the detailed boundary information for fine-grained pavement segmentation.\footnote{Code is available at: \url{https://github.com/yujiaz2/tbnet}.}
    \item To the best of our knowledge, this is the first work attempting to generate fine-grained disease (as well as non-disease) segmentation outputs with multiple classes for detailed pavement inspection.
    \item Experimental results on a newly collected dataset, which specifically targets this fine-grained pavement disease segmentation task demonstrate the effectiveness of the proposed approach.
\end{itemize}

\section{Approach}
In this section, we first formulate the problem of fine-grained pavement disease segmentation, then describe the proposed network in detail. Let the training image set be $\mathbf{S}=\{\mathbf{I}_n,\mathbf{G}_n\}_{n=1}^N$, where $\mathbf{I}_n=\{I_m^{(n)},m=1,…,|\mathbf{I}_n|\}$ is the original input image, and $\mathbf{G}_n=\{G_m^{(n)},m=1,…,|\mathbf{G}_n|,G_m^{(n)}\in\{1,…,C\}\}$ denotes the ground-truth segmented result for $C$ given classes. Note that we have $C=9$ in this task,  including $8$ disease/non-disease classes plus an additional background class. The goal is to train a pixel-wise classifier that minimizes the difference between the output segmentation map and the ground-truth in order to produce the fine-grained segmentation results.

\subsection{Network Architecture}
The architecture of our TB-Net is shown in Figure~\ref{fig2}. The three streams first capture different types of information, and the output features are then fused to obtain the segmentation map. The additional boundary map is produced in parallel to further refine the segmentation results.

\textbf{Spatial Stream.} Since encoding rich spatial information is critical in this dense prediction task, we follow~\cite{yu2018bisenet} and stack three convolutional blocks for its effectiveness. Specifically, each block contains a convolutional layer using 3$\times$3 filters with a stride of 2, and the number of filters for each layer is [64,128,256]. A batch normalization and a ReLU activation are further applied. This shallow structure and the small stride have the benefits that allow the model to preserve the spatial size of the original image, as well as retaining affluent spatial information.

\textbf{Context Stream.} Capturing contextual dependencies plays an important role in semantic segmentation~\cite{yuan2018ocnet,fu2019dual,yu2018bisenet}, thus we propose a context stream that can model a wide range of contextual relationships over local features in order to achieve better feature representations. Given the input image, it incorporates the multi-scale contexts by utilizing a context-aware attention (CAA) mechanism (detailed in Section~\ref{CAA}). We first make use of the ImageNet-pretrained ResNet-101 v2~\cite{he2016identity} to obtain the features from the last layers of the second and the last residual blocks, respectively. In this way, the middle-level and the high-level feature extraction can be achieved. After applying a 1$\times$1 convolutional layer, the transformed features are then fed into two CAA modules to encode the global contextual information and generate two enhanced features.

\textbf{Boundary Stream.}
Making use of boundary information can enhance the segmentation performance, particularly for thin and small objects~\cite{ding2019boundary,takikawa2019gated}. In order to refine the fine-grained segmentation results, we propose a boundary stream that takes the learned global contextual features as the input and generates boundary predictions. This allows the information to flow from the context to the boundary stream. Specifically, we first utilize a residual block with a short-skip connection~\cite{han2017deep,zhang2018image}. This aims to force the network to pay more attention to the informative features. Inspired by the recent success of gated convolutions that learn dynamic feature selection for each channel and each spatial location on images~\cite{yu2019free,jo2019sc}, we then apply a global-gated convolution module (detailed in Section~\ref{GGC}) that assigns more weights to boundary-related pixels by incorporating the global contexts. The output boundary predictions are obtained by further utilizing two transposed convolutional blocks followed by a sigmoid activation.

\textbf{Feature Fusion.}
In the feature fusion module, different types of features are fused to produce the refined segmentation outputs. We first concatenate and transform the outputs of the spatial and the context streams by a batch normalization and a convolutional layer. The encoded features are then fused with the boundary information. Similar to~\cite{hu2018squeeze}, the learned feature maps are refined by using a reweighting mechanism to generate a weight vector. After that, a matrix multiplication operation is performed between the weight vector and the input features, and an element-wise sum operation is further applied to obtain the learned features. The final features are resized to the original image using bilinear interpolation to produce the segmentation map.

\subsection{Context-Aware Attention}
\label{CAA}
Attention mechanism~\cite{vaswani2017attention,fu2019dual,lin2017structured} allows the input to interact with each other and models the global dependencies. Similar to~\cite{fu2019dual} where the self-attention is utilized to capture the semantic interdependencies between any two positions, we employ context-aware attention modules in the context stream to enrich feature representations.

\begin{figure}[t]
\centerline{\includegraphics[width=0.78\linewidth]{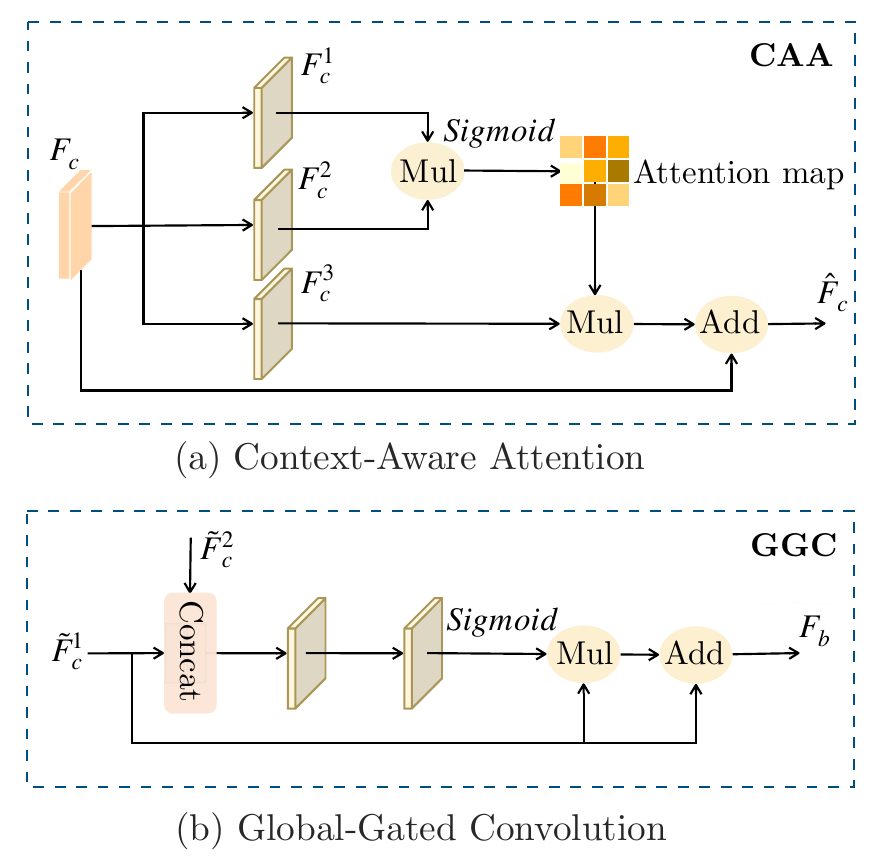}}
\caption{The illustration of the context-aware attention (CAA) module (a) in the context stream and the global-gated convolution (GGC) module (b) in the boundary stream.}
\label{fig3}
\end{figure}

Let the input feature map be $F_c$. As illustrated in Figure~\ref{fig3}(a), three convolutional layers are used to generate the new features $F_c^1$, $F_c^2$ and $F_c^3$. Then the matrix multiplication is utilized on $F_c^1$ and $F_c^2$ after a reshaping operation followed by a sigmoid activation to obtain an attention map. Next, another matrix multiplication is performed between $F_c^3$ and the attention map. The learned context-aware feature $\hat{F}_c$ can be obtained by using an element-wise sum operation with a scale parameter $\gamma$, which is formulated as:
\begin{equation}
\label{eq1}
    \hat{F}_c=\gamma(F_c^3\otimes \sigma(F_c^1\otimes (F_c^2)^T))\oplus F_c,
\end{equation}
where $\sigma$ is the nonlinear activation function, and $\otimes$ and $\oplus$ denote the matrix multiplication and the element-wise sum operation, respectively.

Here the learnable $\gamma$ is initialized as 0, which allows the network to rely on the local cues before assigning more weights to the non-local evidence~\cite{zhang2019self}. In this way, the context-aware attention modules enable the model to capture the affluent relationships by incorporating the semantic interdependencies of the inputs. Note that since there are two context-aware attention modules used in the context stream as shown in Figure~\ref{fig2}, the learnable scale parameters in these two modules are defined as $\gamma^1$ and $\gamma^2$ separately.

\subsection{Global-Gated Convolution}
\label{GGC}
Inspired by~\cite{takikawa2019gated} where a gated mechanism is proposed to ensure that only the boundary-relevant information is processed, we introduce the global-gated convolution module. This aims to highlight the relevant information and filter out the rest to better generate the boundary representations and further improve the segmentation performance. The architecture of the GGC module is shown in Figure~\ref{fig3}(b).

Specifically, let the two input features be $\Tilde{F}_c^1$ and $\Tilde{F}_c^2$. We first concatenate these two features, then perform a batch normalization to balance the different scales of the input features. Next, two 3$\times$3 convolutional layers using 512 filters are applied with a ReLU in between, followed by another batch normalization and a sigmoid activation. After that we multiply the results with the input feature $\Tilde{F}_c^1$ and perform an element-wise sum operation to produce the boundary features $F_b$. In this way, the model can learn to give the boundary regions more weights. The overall process can be summarized as:
\begin{equation}
    F_b=\Tilde{F}_c^1\otimes\sigma(\delta(\Tilde{F}_c^1||\Tilde{F}_c^2))\oplus\Tilde{F}_c^1,
\end{equation}
where $\delta$ denotes the normalized 1$\times$1 convolutional layers and $||$ is the concatenation operation. $\sigma$, $\otimes$ and $\oplus$ are the same as in Equation \eqref{eq1}.

\subsection{Dual-Task Loss Function}
The proposed TB-Net is trained using a dual-task loss on the predicted segmentation and boundary maps. For the segmentation prediction, a cross-entropy loss with a softmax activation is applied on the predicted outputs of the feature fusion module. Due to the fact that pixels in the minority classes tend to be less well trained, we further utilize a weighting mechanism for the highly imbalanced fine-grained disease data, which can be written as:
\begin{equation}
    \mathcal{L}_{WCE}=-\sum_{n=1}^N\omega_n\sum_{c=1}^C\sum_{m=1}^{|\mathbf{I}_n|}y_{m,c}^{(n)}\cdot \log(p_{m,c}^{(n)}),
\end{equation}
where $\mathcal{L}_{WCE}$ is the loss on the predicted segmentation map, $N$ denotes the total number of the training images, and $|\mathbf{I}_n|$ is the number of pixels in each sample. $y_{m,c}^{(n)}$ is the label indicating whether the given pixel in position $m$ of the image $n$ belongs to class $c$ out of the total $C$ classes, and $p_{m,c}^{(n)}$ is the softmax probability of the corresponding predicted output of the network. The weight $\omega_n$ for each image can be defined as:
\begin{equation}
    \omega_n=\sum_{c=1}^C\sum_{m=1}^{|\mathbf{I}_n|}\omega_c\cdot y_{m,c}^{(n)},
\end{equation}
where $\omega_c$ is the class weight which can be computed over all the training images based on the number of pixels for each class by:
\begin{equation}
    \Tilde{\omega}_c=\sum_{c=1}^C\sum_{n=1}^N\sum_{m=1}^{|\mathbf{I}_n|}y_{m,c}^{(n)}\Big/\sum_{n=1}^N\sum_{m=1}^{|\mathbf{I}_n|}y_{m,c}^{(n)}.
\end{equation}
We then normalize the weight by the sum of $\Tilde{\omega}_c$ and obtain the final class weight $\omega_c$. In this way, the weighting mechanism penalizes the loss based on the estimated labels. This is beneficial to reduce the class imbalance that is inherent in the data and avoid ignoring the minority classes.

For the boundary prediction, a binary cross-entropy loss is utilized by introducing the ground-truth boundary information. Here we make use of a recent state-of-the-art method on edge detection~\cite{he2019bi} and generate binary ground-truth boundaries given the annotated segmentation maps. For each image, let $\bar{y}_m^{(n)}\in \{0,1\}$ be the pixel-wise label indicating whether it is a boundary pixel, and $\bar{p}_m^{(n)}$ is the corresponding prediction. The binary cross-entropy loss $\mathcal{L}_{BCE}$ on the predicted maps can be written as:
\begin{equation}
    \mathcal{L}_{BCE}=-\sum_{n=1}^N\sum_{m=1}^{|\mathbf{I}_n|}(\bar{y}_m^{(n)}\cdot \log(\bar{p}_m^{(n)})+(1-\bar{y}_m^{(n)})\cdot \log(1-\bar{p}_m^{(n)})).
\end{equation}
The segmentation and the boundary predictions are jointly learned using the proposed dual-task loss $\mathcal{L}$ as follows:
\begin{equation}
    \mathcal{L}=\lambda_1\mathcal{L}_{WCE}+\lambda_2\mathcal{L}_{BCE},
\end{equation}
where $\lambda_1$ and $\lambda_2$ are the balancing parameters.

\begin{table*}[!b]\centering
\begin{tabular}{c|C{1.1cm}C{1.1cm}C{1.1cm}C{1.1cm}C{1.1cm}C{1.1cm}C{1.1cm}C{1.1cm}|c}
\toprule
\bf{Method} & crack & \makecell[c]{corner-\\fracture} & \makecell[c]{seam-\\broken} & patch & repair & slab & track & light & \bf{Mean} \\
\hline
DeepLabv3~\cite{chen2017rethinking} & 3.43 & 1.08	& 1.42	& \bf{47.37} & 34.93	& 45.81	& 14.55	& 22.25 & 21.35 \\
DeepLabv3+~\cite{chen2018encoder} & 0.33	& 0.78	& 0.91	& 27.89	& \bf{86.51}	& 50.74	& 22.01	& 52.45 & 30.20 \\
DenseASPP~\cite{yang2018denseaspp} & 3.65	& 0.25	& 3.21	& 10.32	& 34.48	& 41.95	& 37.88	& 42.70 & 21.80 \\
BiSeNet+~\cite{yu2018bisenet} & 2.92	& 0.00	& 0.53	& 2.21	& 44.77	& 56.94	& 32.42	& 84.09 & 27.99 \\
BiSeNet~\cite{yu2018bisenet} & \bf{52.60} & 11.74 & 14.10 & 14.00 & 65.40 & \bf{87.52} & 43.29 & 32.86 & 40.19\\
\hline
TB-Net (ours) & 46.34 & \bf{18.98} & \bf{17.65} & 37.15 & 73.98 & 87.36 & \bf{56.78} & \bf{96.02} & \bf{54.28} \\
\bottomrule
\end{tabular}
\caption{Comparison of different methods for fine-grained pavement disease segmentation on our testing set in terms of Class Pixel Accuracy (CPA). The best results are highlighted in bold.}\label{tab1}
\end{table*}

\begin{table*}[!b]\centering
\begin{tabular}{c|C{1.1cm}C{1.1cm}C{1.1cm}C{1.1cm}C{1.1cm}C{1.1cm}C{1.1cm}C{1.1cm}|c}
\toprule
\bf{Method} & crack & \makecell[c]{corner-\\fracture} & \makecell[c]{seam-\\broken} & patch & repair & slab & track & light & \bf{Mean} \\
\hline
DeepLabv3~\cite{chen2017rethinking} & 1.22 & 1.05 & 0.96 & 24.93 & 25.41 & 15.78 & 10.75 & 21.14 & 12.65 \\
DeepLabv3+~\cite{chen2018encoder} & 0.24 & 0.74 & 0.75 & 25.42 & 40.89 & 29.52 & 18.35 & 49.62 & 20.69 \\
DenseASPP~\cite{yang2018denseaspp} & 1.37 & 0.25 & 2.01 & 9.52 & 27.18 & 28.68 & 26.71 & 40.26 & 17.00 \\
BiSeNet+~\cite{yu2018bisenet} & 1.72 & 0.09 & 0.42 & 2.21 & 38.55 & 39.52 & 27.41 & 65.59 & 21.93 \\
BiSeNet~\cite{yu2018bisenet} & 7.59 & 10.15 & 7.56 & 13.63 & 42.57 & 42.16 & 31.28 & 31.03 & 23.25\\
\hline
TB-Net (ours) & \bf{11.43} & \bf{12.28} & \bf{11.34} & \bf{35.77} & \bf{53.30} & \bf{50.74} & \bf{42.19} & \bf{87.54} & \bf{38.07} \\
\bottomrule
\end{tabular}
\caption{Comparison of different methods for fine-grained pavement disease segmentation on our testing set in terms of Intersection-over-Union (IoU). The best results are highlighted in bold.}\label{tab2}
\end{table*}

\section{Experiments}
In this section, we first introduce the newly collected dataset as well as the evaluation metrics, then provide the details of the implementation before presenting the comprehensive experimental analysis.

\subsection{Dataset}
To evaluate the proposed method, we conduct the experiments on an airport pavement disease dataset which specifically targets this fine-grained task to reflect the road surface condition. The dataset contains 3946 images captured with a gray-scale camera on the complex cement/asphalt pavements under different illumination. The examples can be seen in Figure~\ref{fig1}. In total, there are 3171 images for training, and 793 images for testing, which involve eight semantic classes and one background class with pixel-level labels. We randomly select 20 images from the training set for validation. The size of each image ranges from 640$\times$415 to 1800$\times$900, and the number of the annotated areas in the dataset for each class is: $\sharp$\emph{Crack}: 3586, $\sharp$\emph{Cornerfracture}: 151, $\sharp$\emph{Seambroken}: 557, $\sharp$\emph{Patch}: 312, $\sharp$\emph{Repair}: 893, $\sharp$\emph{Slab}: 3040, $\sharp$\emph{Track}: 3749, $\sharp$\emph{Light}: 58.

\subsection{Evaluation Metrics}
We evaluate our performance using two commonly used metrics~\cite{long2015fully} in semantic segmentation: Class Pixel Accuracy (CPA) and Intersection-over-Union (IoU) which are defined as:
\begin{align}
    \begin{split}
        &CPA=\frac{p_{ii}}{\sum_{j=1}^C p_{ij}},\\
        &IoU=\frac{p_{ii}}{\sum_{j=1}^C p_{ij}+\sum_{i=1}^C p_{ji}-p_{ii}},
    \end{split}
\end{align}
where $p_{ij}$ is the number of pixels of class $i$ predicted to belong to class $j$ out of the total $C$ classes. 

\subsection{Implementation Details}
Our TB-Net is implemented using Tensorflow~\cite{abadi2016tensorflow} on a single NVIDIA Tesla V100 16GB GPU. We adopt RMSProp optimizer~\cite{tieleman2017divide} for optimizing the models. In the training stage, the initial learning rate is set to 0.0001, the decay is set to 0.995 and the model gets converged after 150 epochs. The input images and the corresponding ground-truths are resized uniformly to 512$\times$512. In the two context-aware attention modules, since the pre-activation ResNet does not have batch normalization or ReLU in the residual unit output, we further apply a batch normalization and a ReLU on the extracted feature maps. The learned boundary features are upsampled by bilinear interpolation and transformed by a 1$\times$1 convolution before being concatenated in the feature fusion module. We experimentally set the balancing parameters $\lambda_1=\lambda_2=1$. The overall training time is 35 hours and inference is 0.03 seconds.

\subsection{Quantitative Results}
We compare the performance of our TB-Net against four segmentation methods that attain the state-of-the-art results in other semantic segmentation tasks and can be potentially used for this fine-grained pavement segmentation task: (1) DeepLabv3~\cite{chen2017rethinking}, a network employs atrous convolution with upsampled filters to capture the multi-scale contexts; (2) DeepLabv3+~\cite{chen2018encoder}, an encoder-decoder-based model that extends DeepLabv3 by adding a decoder to further refine the segmentation results; (3) DenseASPP~\cite{yang2018denseaspp}, a method concatenates different atrous-convolved features with the multiple scales; (4) BiSeNet~\cite{yu2018bisenet}, a network combines two paths to achieve the rich spatial information and the sizeable receptive field. Note that we keep the learning objective functions the same for fair comparisons.

The results are shown in Table~\ref{tab1} and~\ref{tab2}, where BiSeNet+ and BiSeNet are the same, except that BiSeNet+ does not make use of the weighting mechanism in the loss function. It can be observed that the proposed TB-Net consistently outperforms the existing methods in terms of both mean CPA (mPA) and mean IoU (mIoU). Specifically, for \emph{Cornerfracture} and \emph{Seambroken} where the numbers of the annotated areas are small in the dataset ($\sharp$\emph{Cornerfracture}: 151, $\sharp$\emph{Seambroken}: 557) with each annotated area being relatively small, we still achieve favorable performance over other competing methods. This indicates the importance of fusing the rich spatial and the contextual features with additional boundary information, which can effectively enhance the feature discrimination.

It is worth noting that \emph{Track} contains the markings of various curves and straight lines as well as the low contrast water/oil stains, and our model improves the performance by large margins
(56.78\% vs. 43.29\% in terms of CPA and 42.19\% vs. 31.28\% in terms of IoU). The overall results show that the proposed TB-Net can bring great benefits to fine-grained segmentation across different diseases with varying patterns.

\begin{figure*}[t]
\centerline{\includegraphics[width=0.96\textwidth]{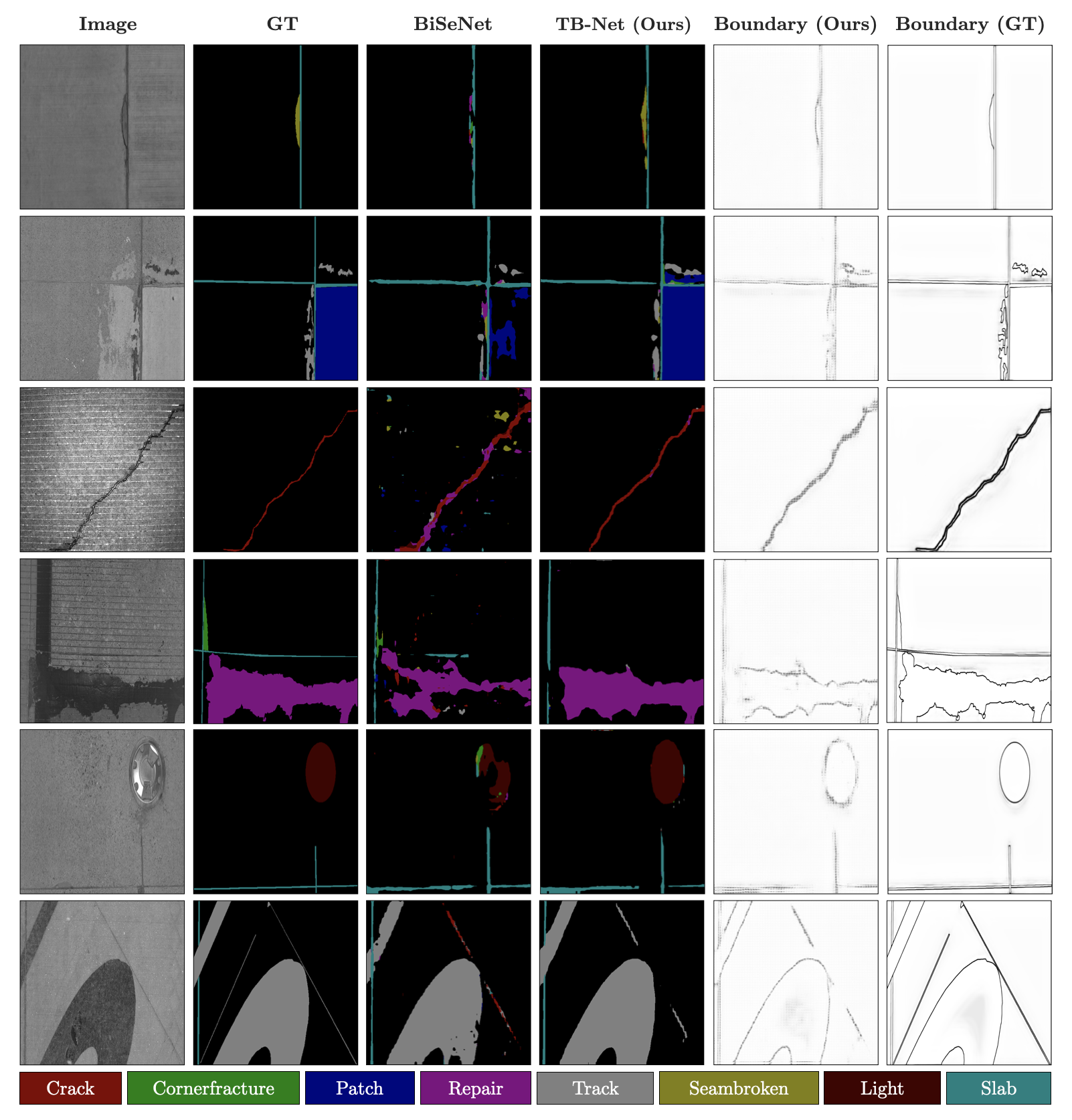}}
\caption{Qualitative results of our proposed TB-Net and one of the competing models, BiSeNet~\cite{yu2018bisenet}. From left to right: image, ground-truth, predictions of BiSeNet and our TB-Net, boundary prediction of TB-Net and boundary ground-truth obtained using~\cite{he2019bi}.}
\label{fig4}
\vspace{-8mm}
\end{figure*}

\begin{table*}[b]\centering
\begin{tabular}{c|c|C{1.1cm}C{1.1cm}C{1.1cm}C{1.1cm}C{1.1cm}C{1.1cm}C{1.1cm}C{1.1cm}|c}
\toprule
\multicolumn{2}{c|}{\bf{Method}} & crack & \makecell[c]{corner-\\fracture} & \makecell[c]{seam-\\broken} & patch & repair & slab & track & light & \bf{Mean} \\
\hline
TB-Net & CPA & 52.40 & 15.88 & 19.17 & 32.10 & 71.89 & 87.02 & 36.63 & 93.73 & 51.10 \\
w/o attn & IoU & 9.08 & 11.58 & 12.41 & 31.16 & 46.01 & 45.33 & 27.64 & 81.80 & 33.12 \\
\hline
TB-Net & CPA & 46.34 & 18.98 & 17.65 & 37.15 & 73.98 & 87.36 & 56.78 & 96.02 & \bf{54.28} \\
(full) & IoU & 11.43 & 12.28 & 11.34 & 35.77 & 53.30 & 50.74 & 42.19 & 87.54 & \bf{38.07} \\
\bottomrule
\end{tabular}
\caption{Results of the ablation experiments for analyzing the effect of the context-aware attention module. We compare our TB-Net to the model that does not make use of the context-aware attention module. The best results are highlighted in bold.}\label{tab3}
\end{table*}

\subsection{Qualitative Results}
We provide visual comparisons of the results obtained by our proposed TB-Net and one of the competing models, BiSeNet~\cite{yu2018bisenet}. We also show the boundaries obtained from the proposed TB-Net and the corresponding ground-truths that are obtained using~\cite{he2019bi}. As shown in Figure~\ref{fig4}, we observe that our model generally achieves better pixel-wise disease segmentation. Specifically, for \emph{Crack} and \emph{Seambroken} where the size of each disease tends to be small, more pixels that belong to these two classes are properly detected by our method. For other diseases where the sizes tend to be large, such as \emph{Patch} and \emph{Repair}, our model can also produce more robust and complete predictions.

\begin{table*}[t]\centering
\begin{tabular}{c|c|C{1.1cm}C{1.1cm}C{1.1cm}C{1.1cm}C{1.1cm}C{1.1cm}C{1.1cm}C{1.1cm}|c}
\toprule
\multicolumn{2}{c|}{\bf{Method}} & crack & \makecell[c]{corner-\\fracture} & \makecell[c]{seam-\\broken} & patch & repair & slab & track & light & \bf{Mean} \\
\hline
TB-Net & CPA & 53.94 & 23.98 & 17.28 & 57.97 & 79.83 & 81.78 & 50.27 & 97.94 & \bf{57.87} \\
w/o BS & IoU & 5.43 & 12.53 & 8.42 & 51.08 & 44.82 & 36.28 & 30.34 & 69.01 & 32.24 \\
\hline
TB-Net & CPA & 46.34 & 18.98 & 17.65 & 37.15 & 73.98 & 87.36 & 56.78 & 96.02 & 54.28 \\
(full) & IoU & 11.43 & 12.28 & 11.34 & 35.77 & 53.30 & 50.74 & 42.19 & 87.54 & \bf{38.07} \\
\bottomrule
\end{tabular}
\caption{Results of the ablation experiments for analyzing the effect of incorporating the boundary information. We compare our TB-Net to the model that does not make use of the boundary information. The best results are highlighted in bold.}\label{tab4}
\end{table*}

\begin{table*}[t]\centering
\begin{tabular}{c|c|C{1.1cm}C{1.1cm}C{1.1cm}C{1.1cm}C{1.1cm}C{1.1cm}C{1.1cm}C{1.1cm}|c}
\toprule
\multicolumn{2}{c|}{\bf{Method}} & crack & \makecell[c]{corner-\\fracture} & \makecell[c]{seam-\\broken} & patch & repair & slab & track & light & \bf{Mean} \\
\hline
TB-Net & CPA & 1.10 & 1.10 & 2.50 & 16.11 & 53.10 & 56.47 & 43.55 & 79.25 & 31.65 \\
w/o w & IoU & 0.95 & 1.10 & 1.87 & 15.72 & 45.29 & 36.37 & 36.70 & 75.02 & 26.63 \\
\hline
TB-Net & CPA & 46.34 & 18.98 & 17.65 & 37.15 & 73.98 & 87.36 & 56.78 & 96.02 & \bf{54.28} \\
(full) & IoU & 11.43 & 12.28 & 11.34 & 35.77 & 53.30 & 50.74 & 42.19 & 87.54 & \bf{38.07} \\
\bottomrule
\end{tabular}
\caption{Results of the ablation experiments for analyzing the effect of the weighting mechanism in the loss function. We compare our TB-Net to the model that does not make use of the weighting mechanism. The best results are highlighted in bold.}\label{tab5}
\end{table*}

Moreover, we can see that the predicted boundaries decently outline the different pavement diseases and help the network produce sharp segmentation results. For example, in the first row in Figure~\ref{fig4}, 
the region for \emph{Seambroken} is rather small and thin which poses more challenges, and BiSeNet fails to predict it correctly. However, our model that incorporates the boundary features better segments this type of disease and produces a higher-quality boundary.

\subsection{Ablation Studies}
\textbf{The effect of the context-aware attention module.} To investigate the use of the context-aware attention in our proposed TB-Net, we drop the attention module in the context stream and compare the performance to the full model. The results are shown in Table~\ref{tab3}. Here we use \emph{TB-Net w/o attn} to denote the model that does not make use of the context-aware attention module. From the table, we observe that the attention module generally helps improve the performance, especially in \emph{Track} with different patterns and sizes of markings and stains, bringing 20.15\% improvements in terms of CPA and 14.55\% in terms of IoU. This indicates that the context-aware attention module can effectively allow the model to incorporate the long-range dependencies between different pixels and further enhance feature representations by capturing the contextual information.

\textbf{The effect of incorporating the boundary information.} We further perform an ablation study to analyze the effect of the boundary information. Table~\ref{tab4} illustrates the performance when dropping the boundary stream (denoted as \emph{TB-Net w/o BS}). Here we fuse the encoded spatial and contextual features and apply a convolutional block, that consists of a 3$\times$3 convolutional layer, a batch normalization and a ReLU. The learned features are then taken as the input to the reweighting mechanism and further predict the segmentation results. We observe that the overall CPA and IoU drop or remain relatively unchanged after dropping the boundary stream for different classes except \emph{Patch}. We hypothesize that this is due to the fact that it is less dependent on the boundaries to segment rectangle-like cement patches. For other small diseases such as \emph{Crack} and \emph{Seambroken} which contain much detailed information, it tends to be more beneficial from fusing the boundary stream.

\textbf{The effect of the weighting mechanism in the loss function.} Since the fine-grained disease data we use is highly imbalanced, we adopt a weighting mechanism in the loss function to avoid the model ignoring the minority classes. The effect of introducing the weighting mechanism can be seen in Table~\ref{tab5}. Here we use \emph{TB-Net w/o w} to denote the model that does not make use of the weighting mechanism. We observe that the full model significantly outperforms the model without the weighting mechanism in terms of both mPA and mIoU. Especially in \emph{Crack} where the number of pixels in each annotated area tends to be small, the performance improves by large margins (45.24\% and 10.48\% in terms of CPA and IoU) by using the weighting mechanism. Besides, the performance of BiSeNet increases compared to BiSeNet+ in Table~\ref{tab1}, which also demonstrates the effectiveness of this weighting mechanism.

\section{Conclusions}
In this work, we address the challenging task of fine-grained pavement disease segmentation and present a three-stream boundary-aware network (TB-Net), that takes full advantages of the low-level spatial, the high-level contextual and the detailed boundary information. The network is trained using a dual-task loss to regularize both the segmentation and the boundary predictions. We evaluate the proposed approach on a newly collected airport pavement disease dataset, and the comprehensive experimental results demonstrate the effectiveness of our TB-Net.
\small
\paragraph{Acknowledgments:}
We are grateful to people at Civil Aviation University of China for collecting, annotating and sharing the data. This work was supported by the National Key Research and Development Project of China (2019YFB1310601), the National Key R\&D Program of China (2017YFC0820203) and the National Natural Science Foundation of China (61673378).

{\small
\bibliographystyle{ieee_fullname}
\bibliography{egbib}

\begin{thebibliography}{10}\itemsep=-1pt

\bibitem{abadi2016tensorflow}
Mart\'{\i}n Abadi, Ashish Agarwal, Paul Barham, Eugene Brevdo, Zhifeng Chen,
  Craig Citro, Greg~S. Corrado, Andy Davis, Jeffrey Dean, Matthieu Devin,
  et~al.
\newblock {TensorFlow}: Large-scale machine learning on heterogeneous systems,
  2015.
\newblock Software available from tensorflow.org.

\bibitem{badrinarayanan2017segnet}
Vijay Badrinarayanan, Alex Kendall, and Roberto Cipolla.
\newblock Segnet: A deep convolutional encoder-decoder architecture for image
  segmentation.
\newblock {\em IEEE Transactions on Pattern Analysis and Machine Intelligence},
  39(12):2481--2495, 2017.

\bibitem{chambon2011automatic}
Sylvie Chambon and Jean-Marc Moliard.
\newblock Automatic road pavement assessment with image processing: Review and
  comparison.
\newblock {\em International Journal of Geophysics}, 2011.

\bibitem{chen2014semantic}
Liang-Chieh Chen, George Papandreou, Iasonas Kokkinos, Kevin Murphy, and Alan~L
  Yuille.
\newblock Semantic image segmentation with deep convolutional nets and fully
  connected crfs.
\newblock {\em arXiv preprint arXiv:1412.7062}, 2014.

\bibitem{chen2017deeplab}
Liang-Chieh Chen, George Papandreou, Iasonas Kokkinos, Kevin Murphy, and Alan~L
  Yuille.
\newblock Deeplab: Semantic image segmentation with deep convolutional nets,
  atrous convolution, and fully connected crfs.
\newblock {\em IEEE Transactions on Pattern Analysis and Machine Intelligence},
  40(4):834--848, 2017.

\bibitem{chen2017rethinking}
Liang-Chieh Chen, George Papandreou, Florian Schroff, and Hartwig Adam.
\newblock Rethinking atrous convolution for semantic image segmentation.
\newblock {\em arXiv preprint arXiv:1706.05587}, 2017.

\bibitem{chen2018encoder}
Liang-Chieh Chen, Yukun Zhu, George Papandreou, Florian Schroff, and Hartwig
  Adam.
\newblock Encoder-decoder with atrous separable convolution for semantic image
  segmentation.
\newblock In {\em Proceedings of the European Conference on Computer Vision},
  pages 801--818, 2018.

\bibitem{ding2019boundary}
Henghui Ding, Xudong Jiang, Ai~Qun Liu, Nadia~Magnenat Thalmann, and Gang Wang.
\newblock Boundary-aware feature propagation for scene segmentation.
\newblock In {\em Proceedings of the IEEE International Conference on Computer
  Vision}, pages 6819--6829, 2019.

\bibitem{dung2019autonomous}
Cao~Vu Dung and Le~Duc Anh.
\newblock Autonomous concrete crack detection using deep fully convolutional
  neural network.
\newblock {\em Automation in Construction}, 99:52--58, 2019.

\bibitem{fan2019road}
Rui Fan and Ming Liu.
\newblock Road damage detection based on unsupervised disparity map
  segmentation.
\newblock {\em IEEE Transactions on Intelligent Transportation Systems}, 2019.

\bibitem{fang2020novel}
Fen Fang, Liyuan Li, Ying Gu, Hongyuan Zhu, and Joo-Hwee Lim.
\newblock A novel hybrid approach for crack detection.
\newblock {\em Pattern Recognition}, page 107474, 2020.

\bibitem{fu2019dual}
Jun Fu, Jing Liu, Haijie Tian, Yong Li, Yongjun Bao, Zhiwei Fang, and Hanqing
  Lu.
\newblock Dual attention network for scene segmentation.
\newblock In {\em Proceedings of the IEEE Conference on Computer Vision and
  Pattern Recognition}, pages 3146--3154, 2019.

\bibitem{goodfellow2014generative}
Ian Goodfellow, Jean Pouget-Abadie, Mehdi Mirza, Bing Xu, David Warde-Farley,
  Sherjil Ozair, Aaron Courville, and Yoshua Bengio.
\newblock Generative adversarial nets.
\newblock In {\em Advances in Neural Information Processing Systems}, pages
  2672--2680, 2014.

\bibitem{han2017deep}
Dongyoon Han, Jiwhan Kim, and Junmo Kim.
\newblock Deep pyramidal residual networks.
\newblock In {\em Proceedings of the IEEE Conference on Computer Vision and
  Pattern Recognition}, pages 5927--5935, 2017.

\bibitem{he2019bi}
Jianzhong He, Shiliang Zhang, Ming Yang, Yanhu Shan, and Tiejun Huang.
\newblock Bi-directional cascade network for perceptual edge detection.
\newblock In {\em Proceedings of the IEEE Conference on Computer Vision and
  Pattern Recognition}, pages 3828--3837, 2019.

\bibitem{he2012guided}
Kaiming He, Jian Sun, and Xiaoou Tang.
\newblock Guided image filtering.
\newblock {\em IEEE Transactions on Pattern Analysis and Machine Intelligence},
  35(6):1397--1409, 2012.

\bibitem{he2016identity}
Kaiming He, Xiangyu Zhang, Shaoqing Ren, and Jian Sun.
\newblock Identity mappings in deep residual networks.
\newblock In {\em Proceedings of the European Conference on Computer Vision},
  pages 630--645, 2016.

\bibitem{hu2018squeeze}
Jie Hu, Li Shen, and Gang Sun.
\newblock Squeeze-and-excitation networks.
\newblock In {\em Proceedings of the IEEE Conference on Computer Vision and
  Pattern Recognition}, pages 7132--7141, 2018.

\bibitem{jo2019sc}
Youngjoo Jo and Jongyoul Park.
\newblock Sc-fegan: Face editing generative adversarial network with user's
  sketch and color.
\newblock In {\em Proceedings of the IEEE International Conference on Computer
  Vision}, pages 1745--1753, 2019.

\bibitem{li2018automatic}
Haifeng Li, Dezhen Song, Yu Liu, and Binbin Li.
\newblock Automatic pavement crack detection by multi-scale image fusion.
\newblock {\em IEEE Transactions on Intelligent Transportation Systems},
  20(6):2025--2036, 2018.

\bibitem{lin2017refinenet}
Guosheng Lin, Anton Milan, Chunhua Shen, and Ian Reid.
\newblock Refinenet: Multi-path refinement networks for high-resolution
  semantic segmentation.
\newblock In {\em Proceedings of the IEEE Conference on Computer Vision and
  Pattern Recognition}, pages 1925--1934, 2017.

\bibitem{lin2017structured}
Zhouhan Lin, Minwei Feng, Cicero Nogueira~dos Santos, Mo Yu, Bing Xiang, Bowen
  Zhou, and Yoshua Bengio.
\newblock A structured self-attentive sentence embedding.
\newblock {\em arXiv preprint arXiv:1703.03130}, 2017.

\bibitem{liu2019deepcrack}
Yahui Liu, Jian Yao, Xiaohu Lu, Renping Xie, and Li Li.
\newblock Deepcrack: A deep hierarchical feature learning architecture for
  crack segmentation.
\newblock {\em Neurocomputing}, 338:139--153, 2019.

\bibitem{long2015fully}
Jonathan Long, Evan Shelhamer, and Trevor Darrell.
\newblock Fully convolutional networks for semantic segmentation.
\newblock In {\em Proceedings of the IEEE Conference on Computer Vision and
  Pattern Recognition}, pages 3431--3440, 2015.

\bibitem{ronneberger2015u}
Olaf Ronneberger, Philipp Fischer, and Thomas Brox.
\newblock U-net: Convolutional networks for biomedical image segmentation.
\newblock In {\em International Conference on Medical Image Computing and
  Computer-Assisted Intervention}, pages 234--241, 2015.

\bibitem{shi2016automatic}
Yong Shi, Limeng Cui, Zhiquan Qi, Fan Meng, and Zhensong Chen.
\newblock Automatic road crack detection using random structured forests.
\newblock {\em IEEE Transactions on Intelligent Transportation Systems},
  17(12):3434--3445, 2016.

\bibitem{takikawa2019gated}
Towaki Takikawa, David Acuna, Varun Jampani, and Sanja Fidler.
\newblock Gated-scnn: Gated shape cnns for semantic segmentation.
\newblock In {\em Proceedings of the IEEE International Conference on Computer
  Vision}, pages 5229--5238, 2019.

\bibitem{tieleman2017divide}
T Tieleman and G Hinton.
\newblock Divide the gradient by a running average of its recent magnitude.
  coursera: Neural networks for machine learning.
\newblock {\em Technical Report.}, 2017.

\bibitem{vaswani2017attention}
Ashish Vaswani, Noam Shazeer, Niki Parmar, Jakob Uszkoreit, Llion Jones,
  Aidan~N Gomez, {\L}ukasz Kaiser, and Illia Polosukhin.
\newblock Attention is all you need.
\newblock In {\em Advances in Neural Information Processing Systems}, pages
  5998--6008, 2017.

\bibitem{xie2020main}
Yu Xie, Fangrui Zhu, and Yanwei Fu.
\newblock Main-secondary network for defect segmentation of textured surface
  images.
\newblock In {\em The IEEE Winter Conference on Applications of Computer
  Vision}, pages 3531--3540, 2020.

\bibitem{yang2018denseaspp}
Maoke Yang, Kun Yu, Chi Zhang, Zhiwei Li, and Kuiyuan Yang.
\newblock Denseaspp for semantic segmentation in street scenes.
\newblock In {\em Proceedings of the IEEE Conference on Computer Vision and
  Pattern Recognition}, pages 3684--3692, 2018.

\bibitem{yu2018bisenet}
Changqian Yu, Jingbo Wang, Chao Peng, Changxin Gao, Gang Yu, and Nong Sang.
\newblock Bisenet: Bilateral segmentation network for real-time semantic
  segmentation.
\newblock In {\em Proceedings of the European Conference on Computer Vision},
  pages 325--341, 2018.

\bibitem{yu2019free}
Jiahui Yu, Zhe Lin, Jimei Yang, Xiaohui Shen, Xin Lu, and Thomas~S Huang.
\newblock Free-form image inpainting with gated convolution.
\newblock In {\em Proceedings of the IEEE International Conference on Computer
  Vision}, pages 4471--4480, 2019.

\bibitem{yuan2018ocnet}
Yuhui Yuan and Jingdong Wang.
\newblock Ocnet: Object context network for scene parsing.
\newblock {\em arXiv preprint arXiv:1809.00916}, 2018.

\bibitem{zhang2019self}
Han Zhang, Ian Goodfellow, Dimitris Metaxas, and Augustus Odena.
\newblock Self-attention generative adversarial networks.
\newblock In {\em International Conference on Machine Learning}, pages
  7354--7363, 2019.

\bibitem{zhang2020crackgan}
Kaige Zhang, Yingtao Zhang, and Heng-Da Cheng.
\newblock Crackgan: Pavement crack detection using partially accurate ground
  truths based on generative adversarial learning.
\newblock {\em IEEE Transactions on Intelligent Transportation Systems}, 2020.

\bibitem{zhang2016road}
Lei Zhang, Fan Yang, Yimin~Daniel Zhang, and Ying~Julie Zhu.
\newblock Road crack detection using deep convolutional neural network.
\newblock In {\em IEEE International Conference on Image Processing}, pages
  3708--3712. IEEE, 2016.

\bibitem{zhang2018image}
Yulun Zhang, Kunpeng Li, Kai Li, Lichen Wang, Bineng Zhong, and Yun Fu.
\newblock Image super-resolution using very deep residual channel attention
  networks.
\newblock In {\em Proceedings of the European Conference on Computer Vision},
  pages 286--301, 2018.

\bibitem{zhu2020weakly}
Jinsong Zhu and Jinbo Song.
\newblock Weakly supervised network based intelligent identification of cracks
  in asphalt concrete bridge deck.
\newblock {\em Alexandria Engineering Journal}, 2020.

\bibitem{zou2018deepcrack}
Qin Zou, Zheng Zhang, Qingquan Li, Xianbiao Qi, Qian Wang, and Song Wang.
\newblock Deepcrack: Learning hierarchical convolutional features for crack
  detection.
\newblock {\em IEEE Transactions on Image Processing}, 28(3):1498--1512, 2018.

\end{thebibliography}
}

\end{document}